\documentclass{article}

\usepackage{natbib}

\usepackage[utf8]{inputenc} 
\usepackage[T1]{fontenc}    
\usepackage{hyperref}       
\usepackage{url}            
\usepackage{booktabs}       
\usepackage{amsfonts}       
\usepackage{nicefrac}       
\usepackage{microtype}      
\usepackage{tikz}
\usepackage{float}

\usepackage{enumitem}
\setlist{nosep} 

\usepackage{mathtools}
\usepackage{amsfonts}       
\usepackage{bbm}
\usepackage{amsthm}
\newtheorem{thm}{Theorem}

\newtheorem{lem}{Lemma}

\newtheorem{prop}{Proposition}
\newtheorem{cor}{Corollary}

\newtheorem{claim}{Claim}
\newtheorem*{rmk}{Remark}

\newtheorem{defn}{Definition}

\usepackage{algorithm}
\usepackage{algpascal}
\usepackage{algorithmicx}
\usepackage{algpseudocode}
\usepackage{tabularx}
\algdef{SE}[SUBALG]{Indent}{EndIndent}{}{\algorithmicend\ }%
\algtext*{Indent}
\algtext*{EndIndent}

\newcommand{\vect}[1]{\boldsymbol{#1}}
\DeclareMathOperator*{\argmax}{arg\,max}

\newcommand\numberthis{\addtocounter{equation}{1}\tag{\theequation}}

\usepackage{xcolor}
\newcount\comments  
\newcommand{\genComment}[2]{\ifnum\comments=1{\textcolor{#1}{\textsf{\footnotesize #2}}}\fi}

\title{Reinforcement Learning in Factored MDPs: Oracle-Efficient Algorithms and Tighter Regret Bounds for the Non-Episodic Setting}

\author{Ziping Xu, Ambuj Tewari}

\begin{document}

\maketitle

\begin{abstract}
We study reinforcement learning in non-episodic factored Markov decision processes (FMDPs). We propose two near-optimal and oracle-efficient algorithms for FMDPs. Assuming oracle access to an FMDP planner, they enjoy a Bayesian and a frequentist regret bound respectively, both of which reduce to the near-optimal bound $\widetilde{O}(DS\sqrt{AT})$ for standard non-factored MDPs. We propose a tighter connectivity measure, factored span, for FMDPs and prove a lower bound that depends on the factored span rather than the diameter $D$. In order to decrease the gap between lower and upper bounds, we propose an adaptation of the REGAL.C algorithm whose regret bound depends on the factored span.
Our oracle-efficient algorithms outperform previously proposed near-optimal algorithms on computer network administration simulations.
\end{abstract}
\section{Introduction}


Designing computationally and statistically efficient algorithms is a core problem in Reinforcement Learning (RL). There is a rich line of works that achieve a strong sample efficiency guarantee with regret analysis in tabular MDPs, where state and action spaces are finite and small \citep{jaksch2010near,osband2013more,dann2015sample,kearns2002near}. A current challenge in RL is dealing with large state and action spaces where even polynomial dependence of regret on state and action spaces size is unacceptable. One idea to meet this challenge is to consider MDPs with compact representations. For example, factored MDPs (FMDPs) \citep{boutilier2000stochastic} represent transition functions of MDPs using a compact Dynamic Bayesian network (DBN) \citep{ghahramani1997learning}. FMDPs have a variety of applications in important real-world problems, e.g. multi-agent RL, and they also serve as important case studies in theoretical RL research \citep{guestrin2002multiagent,guestrin2002context,tavakol2014factored,sun2019model}. 

There is no FMDP planner that is both computationally efficient and accurate \citep{goldsmith1997complexity,littman1997probabilistic}. \citet{guestrin2003efficient} proposed approximate algorithms with prespecified basis functions and bounded approximation errors. For the even harder online learning setting, we study {\it oracle-efficient algorithms}, which learns an unknown FMDP efficiently by assuming an efficient planning oracle. In this paper, our goal is to design efficient online algorithms that only make a polynomial number of calls to the oracle planning oracle. Side-stepping the computational intractability of the offline problem by assuming oracle access to a solver has yielded insights into simpler decision making problems. For example, oracle-based efficient algorithms have been proposed for the contextual bandit problem \citep{syrgkanis2016improved,luo2018efficient}.

Online learning in {\it episodic} FMDP has been studied by \citet{osband2014near}. They proposed two algorithms, PSRL (Posterior Sampling RL) and UCRL-factored with near-optimal Bayesian and frequentist regret bounds, respectively. However, their UCRL-factored algorithm relies on solving a Bounded FMDP \citep{givan2000bounded}, which is an even stronger assumption than the access to a planning oracle.

This work studies FMPDs in the more challenging {\it non-episodic} setting. Previous studies in non-episodic FMDPs either have some high order terms in their analysis \citep{strehl2007model} or depend on some strong connectivity assumptions, e.g. mixing time \citep{kearns1999efficient}. There is no near-optimal regret analysis in this setting yet.

Regret analysis in the non-episodic setting relies on the connectivity assumptions. Previous available connectivity assumptions include {\it mixing time} \citep{lewis2001probabilistic}, {\it diameter} \citep{jaksch2010near} and {\it span of bias vector} \citep{bartlett2009regal}. Mixing time is the strongest assumption and span of bias vector gives the tightest regret bound among the three. However, we show that even upper bound using span can be loose if the factor structure is not taken into account.

This paper makes three main contributions:
\begin{enumerate}
    \item We provide two oracle-efficient algorithms, DORL (Discrete Optimism RL) and PSRL (Posterior Sampling RL), with near-optimal frequentist regret bound and Bayesian regret bound respectively. Both upper bounds depend on the {\em diameter} of the unknown FMDP. The algorithms call the FMDP planner only a polynomial number of times. The upper bound of DORL, when specialized to the standard non-factored MDP setting, matches that of UCRL2 \citep{jaksch2010near}. The same applies to the upper bound of PSRL  in the non-factored setting \citep{ouyang2017learning}. 
    \item We propose a tighter connectivity measure especially designed for FMDPs, called {\it factored span} and prove a regret lower bound that depends on the {\em factored span} of the unknown FMDP rather than its {\em diameter}.
    \item Our last algorithm FSRL is not oracle efficient but its regret scales with factored span, and using it, we are able to {\em reduce the gap between upper and lower bounds on regret} in terms of both the dependence on diameter and on $m$, the number of factors.
\end{enumerate}

\section{Preliminaries}

We first introduce necessary definitions and notation for non-episodic MDPs and FMDPs.

\subsection{Non-episodic MDP}
We consider a setting where a learning agent interacts without resets or episodes with a Markov decision process (MDP), represented by $M = \{\mathcal{S}, \mathcal{A}, P, R\}$, with finite state space $\mathcal{S}$, finite action space $\mathcal{A}$, 
the transition probability $P \in \mathcal{P}_{\mathcal{S}\times\mathcal{A}, \mathcal{S}}$ and reward distribution $R:\mathcal{P}_{\mathcal{S}\times\mathcal{A}, [0, 1]}$. Here $\Delta(\mathcal{X})$ denotes a distribution over the space $\mathcal{X}$. Let $\mathcal{G}(\mathcal{X})$ be the space of all possible distributions over $\mathcal{X}$ and $\mathcal{P}_{\mathcal{X}_1, \mathcal{X}_2}$ is the class of all the mappings from space $\mathcal{X}_1$ to $\mathcal{G}(\mathcal{X}_2)$. Let $S \coloneqq |\mathcal{S}|$ and $A \coloneqq |\mathcal{A}|$.

An MDP $M$ and a learning algorithm $\mathcal{L}$ 
operating on $M$ with an arbitrary initial state $s_1 \in \mathcal{S}$ constitute a stochastic process described by the state $s_t$ visited at time step $t$, the action $a_t$ chosen by $\mathcal{L}$ at step $t$, the reward $r_t \sim R(s_t, a_t)$ and the next state $s_{t+1} \sim P(s_t, a_t)$ obtained for $t = 1, \dots, T$. Let $H_t = \{s_1, a_1, r_1, \dots, s_{t-1}, a_{t-1}, r_{t-1}\}$ be the trajectory up to time $t$.

Below we will define our regret measure in terms of undiscounted sum of rewards. To derive non-trivial upper bounds, we need some connectivity constraint. There are several subclasses of MDPs corresponding to different types of connectivity constraints (e.g., see the discussion in \citet{bartlett2009regal}). We first focus on the class of {\em communicating} MDPs, i.e., the diameter of the MDP, which is defined below, is upper bounded by some $D < \infty$.

\begin{defn}[Diameter]
Consider the stochastic process defined by a stationary policy $\pi:\mathcal{S} \rightarrow \mathcal{A}$ operating on an MDP $M$ with initial state $s$. Let $T(s^{\prime} \mid M, \pi, s)$ be the random variable for the first time step in which state $s^{\prime}$ is reached in this process. Then the diameter of $M$ is defined as
$$
D(M) :=\max_{s \neq s^{\prime} \in \mathcal{S}} \min_{\pi : \mathcal{S} \rightarrow \mathcal{A}} \mathbb{E}\left[T\left(s^{\prime} | M, \pi, s\right)\right].
$$
\end{defn}

A stationary policy $\pi$ on an MDP $M$ is a mapping $\mathcal{S} \mapsto \mathcal{A}$.
An average reward (also called gain) of a policy $\pi$ on $M$ with an initial distribution $s_1$ is defined as
$$
    \lambda(M, \pi, s_1) = \limsup_{T \rightarrow \infty} \frac{1}{T} \mathbb{E}[\sum_{t = 1}^T r(s_t, \pi(s_t))],
$$ where the expectation is over trajectories $H_T$. We restrict the choice of policies within the set $\Pi$ of all policies whose average reward is independent of the starting state $s_1$. It can be shown that for a communicating MDP the optimal policies with the highest average reward are in the set and neither the optimal policy nor the optimal reward depends on the initial state. 
Let $\pi(M) = \argmax_{\pi \in \Pi} \lambda(M, \pi, s_1)$ denote the optimal policy for MDP $M$ and $\lambda^*(M)$ denote the optimal average reward or optimal gain. For any optimal policy $\pi(M)$, following the definition in \cite{puterman2014markov}, we define 
$$
    h(M, s) = \mathbb{E}[ \sum_{t = 1}^{\infty} (r_t - \lambda^{*}(M)) \mid s_1 = s],\text{ for } s = 1, \dots S,
$$
where the expectation is taken over the trajectory generated by policy $\pi(M)$. And the bias vector of MPD $M$ is $\vect{h}(M) \coloneqq (h(M, 1), \dots, h(M, S))^T$. Let the span of a vector $\vect{h}$ be $sp(\vect{h}) \coloneqq \max_{s_1, s_2} \vect{h}(s_1) - \vect{h}(s_2)$. Note that if there are multiple optimal policies, we consider the policy with the largest span for its bias vector.

We define the regret of a reinforcement learning algorithm $\mathcal{L}$ operating on MDP $M$ up to time $T$ as
$$
    R_T \coloneqq \sum_{t=1}^{T}\left(\lambda^*(M)-r_{t}\right),
$$
and Bayesian regret w.r.t. a prior distribution $\phi$ on a set of MDPs as $\mathbb{E}_{M \sim \phi} R_T$.


\paragraph{Optimality equation for average reward criterion.} 
We let $\vect{R}(M, \pi)$ denote the $S$-dimensional vector with each element representing $\mathbb{E}_{r\sim R(s, \pi(s))}[r]$ and $P(M, \pi)$ denote the $S\times S$ matrix with each row as $P(s, \pi(s))$.
For any communicating MDP $M$, the bias vector $\vect{h}(M)$ satisfies the following equation \citep{puterman2014markov}:
\begin{equation}
     \vect{1} \lambda^*(M) + \vect{h}(M) = \vect{R}(M, \pi^*) + P(M, \pi^*)\vect{h}(M). \label{equ:biasVector}
\end{equation}



\subsection{Factored MDP}
Factored MDP is modeled with a DBN (Dynamic Bayesian Network) \citep{dean1989model}, where transition dynamics and rewards are factored and each factor only depends on a finite scope of state and action spaces. We use the definition in \citet{osband2014near}. We call $\mathcal{X} = \mathcal{S} \times \mathcal{A}$ factored set if it can be factored by $\mathcal{X}=\mathcal{X}_{1} \times \ldots \times \mathcal{X}_{n}$. Note this formulation generalizes those in \cite{strehl2007model,kearns1999efficient} to allow the factorization of the action set as well.

\begin{defn} [Scope operation for factored sets]
For any subset of indices $Z \subseteq\{1,2, . ., n\}$, let us define the scope set $\mathcal{X}[Z]:=\bigotimes_{i \in Z} \mathcal{X}_{i}$. Further, for any $x \in \mathcal{X}$ define the scope variable $x[Z] \in \mathcal{X}[Z]$ to be the value of the variables $x_i \in \mathcal{X}_i$ with indices $i \in Z$. For singleton sets $\{i\}$, we write $x[i]$ for $x[\{i\}]$ in the natural way.
\end{defn}{}

\begin{defn}[Factored reward distribution]
A reward distribution $R$ is factored over $\mathcal{X}$ with scopes $Z^R_1, \dots, Z^R_l$ if and only if, for all $x \in \mathcal{X}$, there exists distributions $\left\{R_{i} \in \mathcal{P}_{\mathcal{X}\left[Z^P_{i}\right], [0, 1]}\right\}_{i=1}^{l}$ such that any $r \sim R(x)$ can be decomposed as $\sum_{i = 1}^l r_i$, with each $r_i \sim R_i(x[Z^R_i])$ individually observable. Throughout the paper, we also let $R(x)$ denote reward function of the distribution $R(x)$, which is the expectation $\mathbb{E}_{r \sim R(x)}[r]$.
\end{defn}

\begin{defn}[Factored transition probability]
A transition function $P$ is factored over $\mathcal{S} \times \mathcal{A}=\mathcal{X}_{1} \times \ldots \times \mathcal{X}_{n}$ and $\mathcal{S} = \mathcal{S}_{1} \times \ldots \mathcal{S}_{m}$  with scopes $Z^P_{1}, \ldots, Z^P_{m}$ if and only if, for all $x \in \mathcal{X}, s \in \mathcal{S}$ there exist some $\left\{P_{i} \in \mathcal{P}_{\mathcal{X}\left[Z_{i}\right], \mathcal{S}_{i}} \right\}_{i=1}^{m}$ such that,
$$
    P(s | x)=\prod_{i=1}^{m} P_{i}\left(s[i] \mid x\left[Z_{i}\right]\right).
$$
For simplicity, let $P(x)$ also denote the vector for the probability of each next state from current pair $x$. We define $P_i(x)$ in the same way.
\end{defn}{}

\paragraph{Assumptions on FMDP.} To ensure a finite number of parameters, we assume that $|\mathcal{X}[Z_i^R]| \leq L$ for $i \in [n]$, $|\mathcal{X}[Z_i^P]| \leq L$ for $i \in [m]$ and $|\mathcal{S}_i| \leq W$ for all $i \in [m]$ for some finite $L$ and $W$. Furthermore, we assume that $r \sim R$ is in $[0, 1]$ with probability 1.

\paragraph{Empirical estimates.} We first define number of visits for each factored set. Let $N_{R_i}^t(x) \coloneqq \sum_{\tau = 1}^{t-1}\mathbbm{1}\{x_\tau[Z_i^R] = x\}$ be the number of visits to $x \in \mathcal{X}[Z_i^R]$ until $t$, 
$N_{P_i}^t(x)$ be the number of visits to $x \in \mathcal{X}[Z_i^P]$ until $t$ and $N_{P_i}^t(s, x)$ be the number of visits to $x \in \mathcal{X}[Z_i^P], s \in \mathcal{S}_i$ until $t$. The empirical estimate for $R_i(x)$ is $\hat{R}^t_i(x) = \sum_\tau^{t-1} r_\tau \mathbbm{1}\{x_\tau[Z_i^R] = x\} / \max\{1, N^t_{R_i}(x)\}$ for $i \in [l]$. Estimate for transition probability is $\hat{P}^t_i(s \mid x) = \frac{N_{P_i}^t(s, x)}{\max\{1, N_{P_i}^t(x)\}}$ for $i \in [m]$. We let $N^k_{R_i}, \hat R_i^k$ and $\hat P_i^k$ be $N^{t_k}_{R_i}, \hat R_i^{t_k}$ and $\hat P_i^{t_k}$ with $t_k$ be the first step of episode $k$.

\section{Oracle-efficient Algorithms}
We use PSRL (Posterior Sampling RL) and a modified version of UCRL-factored, called DORL (Discrete Optimism RL). Both PSRL and DORL use a fixed policy within an episode. For PSRL, we apply optimal policy for an MDP sampled from the posterior distribution of the true MDP. For DORL, instead of optimizing over a bounded MDP, we construct a new extended MDP, which is also factored with the number of parameters polynomial in that of the true MDP. Then the optimal policy of the extended FMDP is mapped to the policy space of the true FMDP. Instead of using dynamic episodes, we show that a simple fixed episode scheme can also give us near-optimal regret bounds. Algorithm details are shown in Algorithm \ref{alg:DORL}.


\subsection{Extended FMDP}
\paragraph{Previous two constructions.} Previous near-optimal algorithms on regular MDP depend on constructing an extended MDP with a high probability of being optimistic.
\citet{jaksch2010near} constructs the extended MDP with a continuous action space to allow choosing any transition probability in a confidence set. 
This construction generates a bounded-parameter MDP. \citet{agrawal2017optimistic} instead sample transition probability only from the extreme points of a similar confidence set and combine them by adding extra discrete actions.

Solving the bounded-parameter MDP by the first construction, which requires storing and ordering the $S$-dimensional bias vector, is not feasible for FMDPs. There is no direct adaptation that mitigates this computation issue. We show that the second construction using only a discrete set of MDPs, by removing the sampling part, can be solved with a much lower complexity in the FMDP setting. 

We formally describe the construction. For simplicity, we ignore the notations for $k$ in this section. First define the error bounds as an input. For every $x \in \mathcal{X}[Z_i^P]$, $s \in \mathcal{S}$, we have an error bound $W_{P_i}(s \mid x)$ for transition probability $\hat P_i(s \mid x)$. For every $x \in \mathcal{X}[Z_i^R]$, we have an error bound $W_{R_i}(x)$ for $\hat R_i(x)$. At the start of episode $k$ the construction takes the inputs of $\hat M_k$ and the error bounds, and outputs the extended MDP $M_k$.


\paragraph{Extreme transition dynamic.}
We first define the extreme transition probability mentioned above in factored setting. Let $P_i(x)^{s+}$ be the transition probability that encourages visiting $s \in \mathcal{S}_i$, be
$$
    P_i(x)^{s+} = P_i(x) - W_{P_i}(\cdot \mid x) + \mathbbm{1}_{s} \sum_jW_{P_i}(j \mid x),
$$
where $\mathbbm{1}_{j}$ is the vector with all zeros except for a one on the $j$-th element.
By this definition, $P_i(x)^{s+}$ is a new transition probability that puts all the uncertainty onto the direction $s$. An example is shown in Figure \ref{fig:expl}. Our construction assigns an action for each extreme transition dynamic.

\begin{figure}[H]
    \centering
    $
    \underbrace{
    \left(\begin{array}{c}
    0.5 \\ 0.3 \\ 0.2
    \end{array}\right)
    }_{\text{Estimated dynamic}}
    -
    \underbrace{
    \left(\begin{array}{c}
    0.1 \\ 0.05 \\ 0.05
    \end{array}\right)
    }_{\text{Uncertainty}}
    +
    \underbrace{
    \left(\begin{array}{c}
    0 \\ 0.2 \\ 0
    \end{array}\right)
    }_{\text{Encourage visiting } s_2}
    =
    \underbrace{
    \left(\begin{array}{c}
    0.4 \\ 0.45 \\ 0.15
    \end{array}\right)
    }_{\text{Extreme dynamic}}
    $
    \caption{An extreme transition dynamic that encourages visiting the second state out of three states.}
    \label{fig:expl}
\end{figure}

\paragraph{Construction of extended FMDP.} Our new factored MDP $M_k = \{\mathcal{S}, \tilde{\mathcal{A}}, \tilde P, \tilde R\}$, where $\tilde{\mathcal{A}} = \mathcal{A} \times \mathcal{S}$ and the new scopes $\{\tilde Z_i^R\}_{i = 1}^{l}$ and $\{\tilde Z_i^P\}_{i = 1}^{m}$ are the same as those for the original MDP. 

Let $\tilde{\mathcal{X}} = \mathcal{X} \times {\mathcal{S}}$. The new transition probability is factored over $\tilde{\mathcal{X}} = \bigotimes_{i \in [m]}(\mathcal{X}[Z_i^P] \times \mathcal{S}_i)$ and $\mathcal{S} = \bigotimes_{i \in [m]}\mathcal{S}_i$ with the factored transition probability to be
$$
    \tilde{P}_i(x, s[i]) \coloneqq \hat P_i(x)^{s[i]+}, \text{ for any } x \in \mathcal{X}[Z_i^P], s \in \mathcal{S}.
$$
The new reward function is factored over $\tilde{\mathcal{X}} = \bigotimes_{i \in [l]}(\mathcal{X}[Z_i^P] \times \mathcal{S}_i)$, with reward functions to be
$$
    \tilde R_i(x, s[i]) = \hat R_i(x) + W_{R_i}(x), 
$$
for any $x \in \mathcal{X}[Z_i^R], s \in \mathcal{S}$.

\begin{claim}
    The factored set $\tilde{\mathcal{X}} = \mathcal{S}\times \tilde{\mathcal{A}}$ of the extended MDP $M_k$ satisfies each $|\tilde{\mathcal{X}}[Z_i^P]| \leq LW$ for any $i \in [m]$ and each $|\tilde{\mathcal{X}}[Z_i^R]| \leq LW$ for any $i \in [l]$. \label{lem:size_M_k}
\end{claim}

By Claim \ref{lem:size_M_k}, any planner that efficiently solves the original MDP, can also solve the extended MDP. We find the best policy $\tilde{\pi}_k$ for $M_k$ using the planner. To run a policy $\pi_k$ on original action space, we choose $\pi_k$ such that $(s, \pi_k(s)) = f(s, \tilde{\pi}_k(s))$ for every $s \in \mathcal{S}$, where $f:\tilde{\mathcal{X}} \mapsto \mathcal{X}$ maps any new state-action pair to the pair it is extended from, i.e. $f(x^s) = x$ for any $x^s \in \tilde{\mathcal{X}}$.

\begin{algorithm}[h]
   \caption{PSRL and DORL}
\begin{algorithmic}
   \State {\bfseries Input:} $\mathcal{S}, \mathcal{A}$, accuracy $\rho$ for DORL and prior distribution for PSRL, $T$, encoding $\mathcal{G}$ and $L$, upper bound on the size of each factor set.
   \State $k \leftarrow 1; t \leftarrow 1; t_k \leftarrow 1; T_k \leftarrow 1; \mathcal{H} = \{\}$.
   \Repeat
   \State For DORL:
   \Indent
       \State Construct the extended MDP $M_k$ using error bounds:
       \begin{align*}
       W_{P_i}^k(s \mid x) = \min\{\sqrt{\frac{18 \hat{P}_i(s | x) \log (c_{i, k})}{\max \{N_{P_i}^k(x), 1\}}}+\frac{18 \log (c_{i, k})}{\max \{N_{P_i}^k(x), 1\}}, \hat{P}_i^{k}(s | x)\}, \numberthis \label{equ:delta_p}
       \end{align*}{}
       \State for $c_{i, k} = 6mS_i|\mathcal{X}[Z_i^P]|t_k/\rho$ and 
       \begin{equation}
       W_{R_i}^k = \sqrt{\frac{12\log(6l|\mathcal{X}[Z_i^R]|t_k/\rho)}{\max \{N_{R_i}(x), 1\} }}. \numberthis \label{equ:delta_r}
       \end{equation}
       \State Compute $\tilde \pi_k = \pi(M_k)$ and find corresponding $\pi_k$ in original action space.
   \EndIndent
   \State For PSRL:
   \Indent
       \State Sample $M_k$ from $\phi(M | \mathcal{H})$.
       \State Compute $\pi_k = \pi(M_k)$.
   \EndIndent
   \For{$t=t_k$ {\bfseries to} $t_k + T_k - 1$}
   \State Apply action $a_t = \pi_k(s_t)$
   \State Observe new state $s_{t+1}$.
   \State Observe new rewards $r_{t+1} = (r_{t+1, 1}, \dots r_{t+1, l})$.
   \State $\mathcal{H} = \mathcal{H} \cup \{(s_t, a_t, r_{t+1}, s_{t+1})\}$.
   \EndFor
   \State $k \leftarrow k + 1$.
   \State $T_k \leftarrow \lceil k/L \rceil$; $t_k \leftarrow t+1$.
   \Until{$t_k > T$}
\end{algorithmic}
\label{alg:DORL}
\end{algorithm}

\section{Upper bounds for PSRL and DORL}
We achieve the near-optimal Bayesian regret bound by PSRL and frequentist regret bound by DORL, respectively. Let $\tilde{O}$ denote the order ignoring the logarithmic term and the universal constant.
\begin{thm}[Regret of PSRL]
\label{thm:1}
Let $M$ be the factored MDP with graph structure $\mathcal{G} = \left(\left\{\mathcal{S}_{i}\right\}_{i=1}^{m} ;\left\{\mathcal{X}_{i}\right\}_{i=1}^{n} ;\left\{Z_{i}^{R}\right\}_{i=1}^{l} ;\left\{Z_{i}^{P}\right\}_{i=1}^{m}\right)$,  all $|\mathcal{X}[Z_i^R]|$ and $|\mathcal{X}[Z_j^P]| \leq L$,  $|\mathcal{S}_i| \leq W$ and diameter upper bounded by $D$. 
Then if $\phi$ is the true prior distribution over the set of MDPs with diameter $\leq$ D, then we bound Bayesian regret of PSRL:
$$
    \mathbbm{E} [R_T]
    = \tilde{O}(D(l +m \sqrt{W}) \sqrt{TL}).
$$
\end{thm}

\begin{thm}[Regret of DORL]
\label{thm:2}
Let $M$ be the factored MDP with graph structure $\mathcal{G} = \left(\left\{\mathcal{S}_{i}\right\}_{i=1}^{m} ;\left\{\mathcal{X}_{i}\right\}_{i=1}^{n} ;\left\{Z_{i}^{R}\right\}_{i=1}^{l} ;\left\{Z_{i}^{P}\right\}_{i=1}^{m}\right)$, all $|\mathcal{X}[Z_i^R]|$ and $|\mathcal{X}[Z_j^P]| \leq L$,  $|\mathcal{S}_i| \leq W$ and diameter upper bounded by $D$. 
Then, with high probability, regret of DORL is upper bounded by:
$$
    R_T
    = \tilde{O}(D(l +m \sqrt{W}) \sqrt{TL}).
$$
\end{thm}

The two bounds match the frequentist regret bound in \citet{jaksch2010near} and Bayesian regret bound in \cite{ouyang2017learning} for non-factored communicating MDP. We also give a condition of designing the speed of changing policies.

\begin{rmk}
Replacing the episode length in Algorithm \ref{alg:DORL} with any $\{T_k\}_{k = 1}^K$ that satisfies $K = O(\sqrt{LT})$ and $T_k = O(\sqrt{T/L})$ for all $k \in [K]$, the frequentist bound in Theorem \ref{thm:2} still holds. Furthermore, if $\{T_k\}_{k = 1}^K$ is fixed the Bayesian bound in Theorem \ref{thm:2} also holds. \label{rmk:1}
\end{rmk}

\section{Lower Bound and Factored Span}

Any regret bound depends on a difficulty measure determining the connectivity of the MDP. The upper bounds of DORL and PSRL use diameter. A tighter alternative is the span of bias vector \citep{bartlett2009regal}, defined as $sp(\vect{h}^*)$, where $\vect{h}^*$ is the bias vector of the optimal policy. However, none of those connectivity measures address the complexity of the graph structure. Indeed, some graph structure allows a tighter regret bound. In this section, we first show a lower bound with a Cartesian product structure. We further propose a new connectivity measure that can scale with the complexity of the graph structure.


\paragraph{Large diameter case.} We consider a simple FMDP with infinite diameter but still solvable. The FMDP is a Cartesian product of two identical MDPs, $M_1$ and $M_2$ with $\mathcal{S} = \{0, 1, 2, 3\}$, $\mathcal{A} = \{1, 2\}$. The transition probability is chosen such that from any state and action pair, the next state will either move forward or move backward with probability one (state 0 is connected with state 3 as a circle).

We can achieve a low regret easily by learning each MDP independently. However, since the sum of the two states always keeps the same parity, vector state $(0, 1)$ can never be reached from $(0, 0)$. Thus, the FMDP has an infinite diameter. The span of bias vector, on the other hand, is upper bounded by $D(M_1) + D(M_2)$, which is tight in this case.

\paragraph{Lower bound with only dependency on span.} Let us formally state the lower bound. Our lower bound casts some restrictions on the scope of transition probability, i.e. the scope contains itself, which we believe is a natural assumption. We provide a proof sketch for Theorem \ref{thm:3} here.

\begin{thm}[Lower bound]
\label{thm:3}
For any algorithm, any graph structure satisfying $\mathcal{G} = \left(\left\{\mathcal{S}_{i}\right\}_{i=1}^{n} ;\left\{\mathcal{S}_i\times \mathcal{A}_i\right\}_{i=1}^{n} ;\left\{Z_{i}^{R}\right\}_{i=1}^{n} ;\left\{Z_{i}^{P}\right\}_{i=1}^{n}\right)$ with $|\mathcal{S}_i| \leq W$, $|\mathcal{X}[Z_i^R]| \leq L$, $|\mathcal{X}[Z_i^P]| \leq L$ and $i \in Z_i^P$ for $i \in [n]$, there exists an FMDP with the span of bias vector $sp(\vect{h}^+)$, such that for any initial state $s \in \mathcal{S}$, the expected regret of the algorithm after $T$ step is 
\begin{equation}
    \Omega(\sqrt{sp(\vect{h^+})LT}). \label{equ:lowerbound}
\end{equation}
\end{thm}

The proof is given in Appendix \ref{app:lowerbound}. As we can see, the upper bound in Theorem \ref{thm:1} is larger than the lower bound by a factor of $\frac{D}{\sqrt{sp(\vect{h}^+)}}$, $m$, $l$ and $\sqrt{W}$. We now discuss how to reduce the first three excesses. 

\subsection{Tighter connectivity measure}
The mismatch in the dependence on $m$ is due to not taking the factor structure into account properly in the definition of the span. A tighter bound should be able to detect the easier structure, e.g. product of independent MDPs. We now propose  {\it factored span} that scales with the complexity of the graph structure.
\begin{defn}[Factored span]
For an FMDP $M$ with an bias vector $\vect{h}$ of it optimal policy and a factorization of state space $\mathcal{S} = \bigotimes_{i=1}^m \mathcal{S}_i$, we define factored span $sp_1, \dots, sp_m$ as:
$$
    sp_i \coloneqq \max_{s_{-i}} sp(\vect{h}(\cdot, s_{-i})) \text{ and let } Q(\vect{h}) \coloneqq \sum_{i = 1}^m sp_i, 
$$
where $s_{-i} \coloneqq (s_1, \dots, s_{i-1}, s_{i+1}, \dots, s_m)$ and $sp(h(\cdot, s_{-i})) \coloneqq (h(s, s_{-i}))_{s \in \mathcal{S}_1}$.
\end{defn}

\begin{prop}
For any bias vector $\vect{h}$, $sp(\vect{h}) \leq Q(\vect{h}) \leq m\ sp(\vect{h})$. The first equality holds when the FMDP has the construction of Cartesian product of $m$ independent MDPs. The lower bound (\ref{equ:lowerbound}) can also be written as $\Omega(\sqrt{Q(\vect{h^+})LT})$. 
\end{prop}

\subsection{Tighter upper bound}

We now provide another algorithm called FSRL (Factored-span RL) with a tighter regret bound of $\tilde{O}(Q(\vect{h})\sqrt{WLT})$ as shown in Theorem \ref{thm:4}. 
The bound reduces the gap on $m$, $l$ and replaces $D$ with the sum of factored span $Q$. Proposition 1 guarantees that $Q(\vect{h}) \leq m sp(\vect{h}) \leq m D$ such that the upper bound is at least as good as the upper bound in Theorem \ref{thm:1}.

FSRL (full description in Appendix~\ref{sec:fsrl}, Algorithm \ref{alg:fregal}), mimics REGAL.C by solving the following optimization, 
   $$
    M = \argmax_{M \in \mathcal{M}_k} \lambda^*(M) \quad \text{subject to} \quad Q(\vect{h}(M)) \leq Q \text{ for some prespecified } Q > 0,
   $$
  where $\mathcal{M}_k$ is the confidence set defined in (\ref{equ:delta_p}) and (\ref{equ:delta_r}).
FSRL relies on the computational oracle of optimizing average rewards over the confidence set with the sum of factored span bounded by a prespecified value. Therefore, FSRL cannot be run by just calling an FMDP planning oracle.

\begin{thm}[Regret of FSRL (Factored-span RL)]
\label{thm:4}
Let $M$ be the factored MDP with graph structure $\mathcal{G} = \left(\left\{\mathcal{S}_{i}\right\}_{i=1}^{m} ;\left\{\mathcal{X}_{i}\right\}_{i=1}^{n} ;\left\{Z_{i}^{R}\right\}_{i=1}^{l} ;\left\{Z_{i}^{P}\right\}_{i=1}^{m}\right)$, all $|\mathcal{X}[Z_i^R]|$ and $|\mathcal{X}[Z_j^P]| \leq L$,  $|\mathcal{S}_i| \leq W$, bias vector of optimal policy $\vect{h}$ and its sum of factored spans $Q(\vect{h})$. 
Then, with high probability, regret of FSRL is upper bounded by:
$
    R_T
    = \tilde{O}(Q(\vect{h})\sqrt{WLT}).
$
\end{thm}

The proof idea is on bounding the deviation of transition probabilities between the true MDP and $M_k$ in episode $k$ with factored span. The details are shown in Appendix \ref{app:thm4}.


\section{Simulation}

\begin{figure}[tb]
\centering
\includegraphics[scale=0.45]{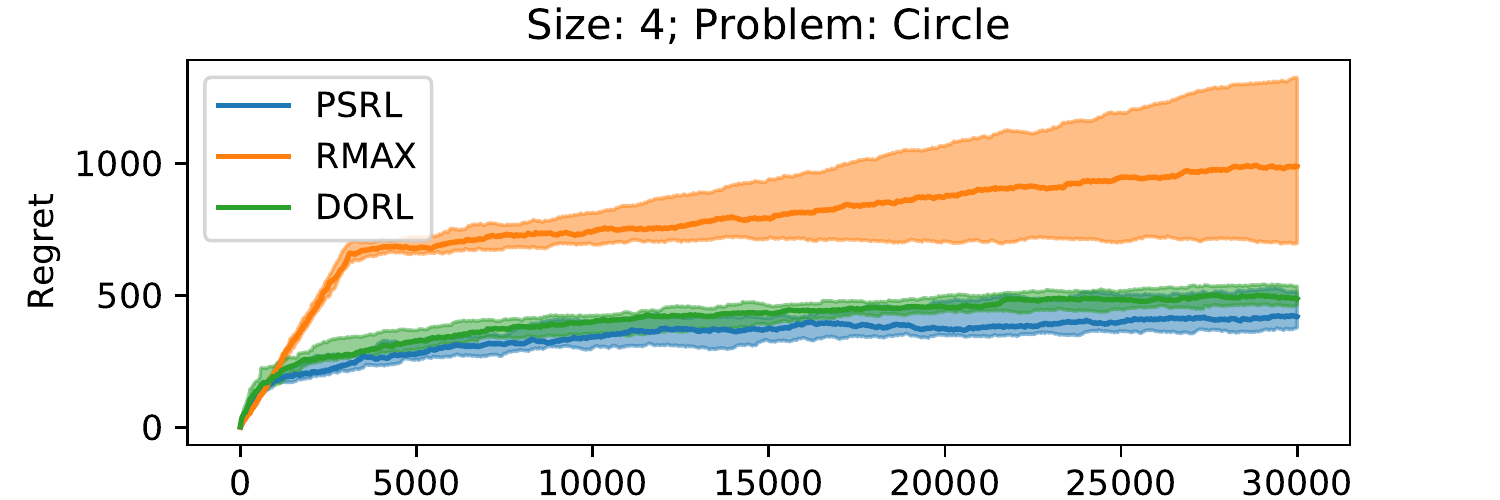}
\includegraphics[scale=0.45]{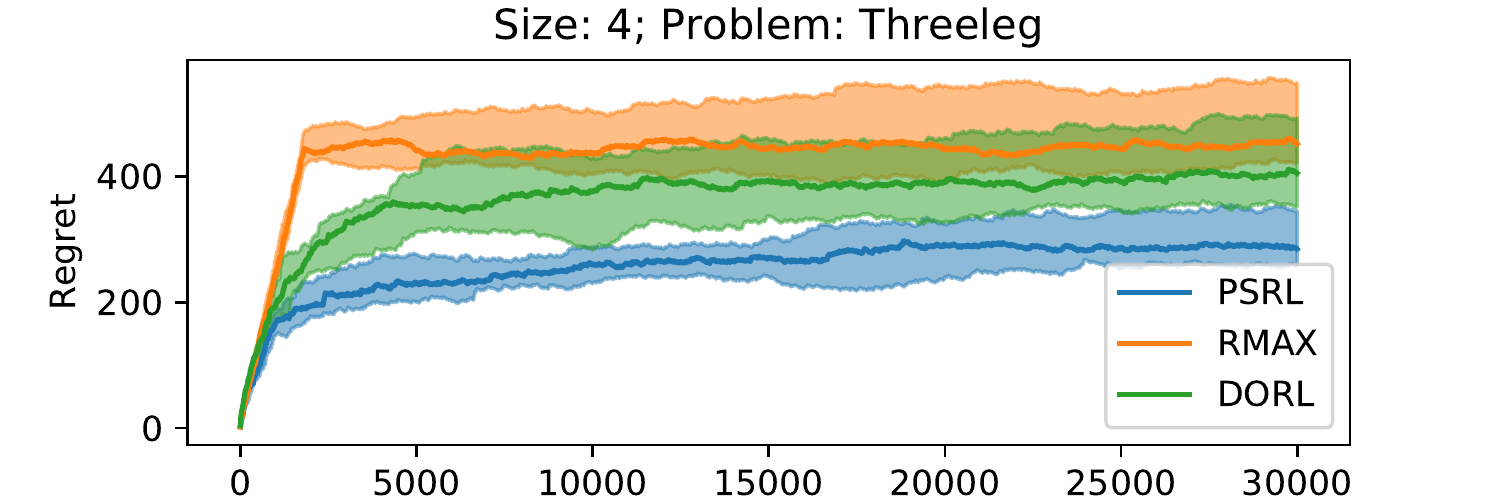}
\includegraphics[scale=0.45]{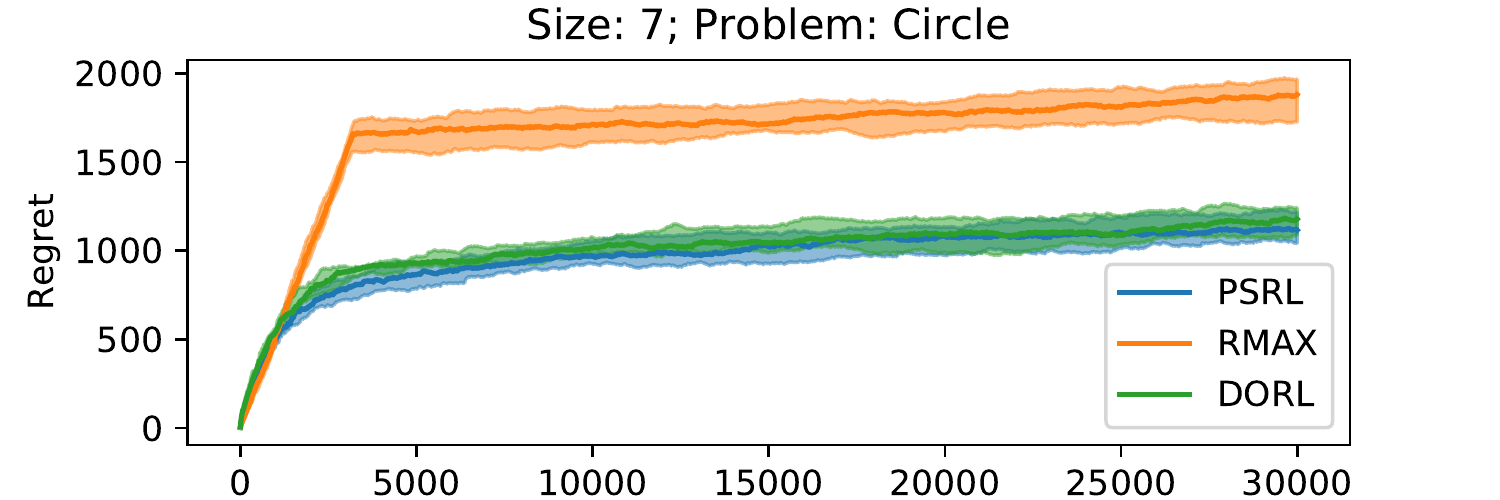}
\includegraphics[scale=0.45]{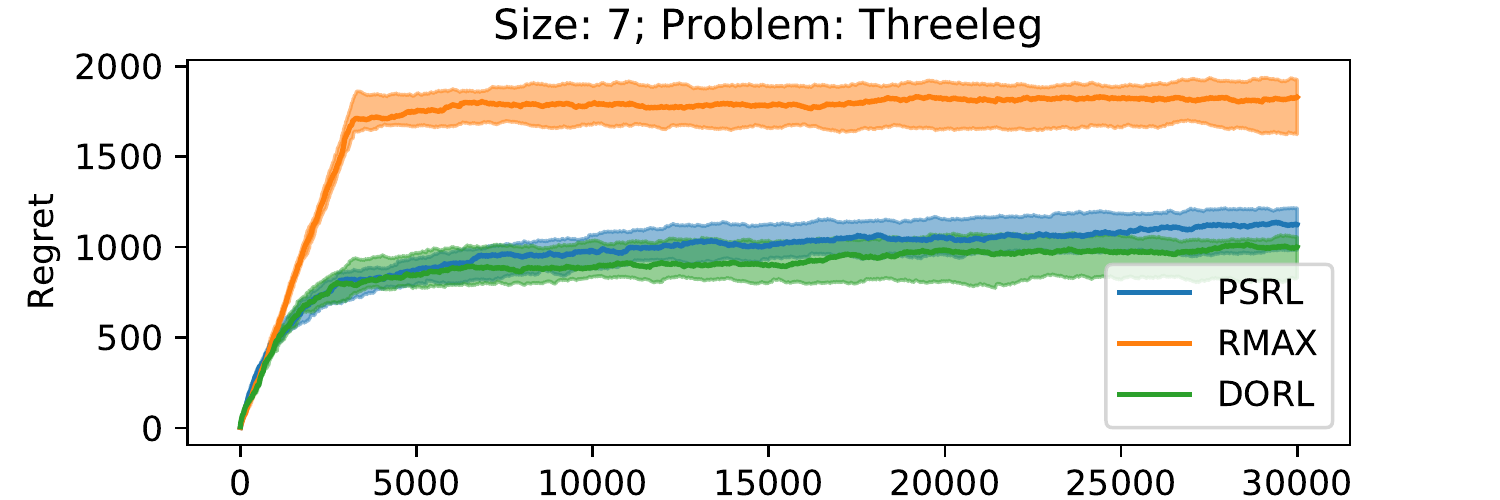}
\caption{Regrets of PSRL, f-Rmax and DORL on circle and three-leg MDP with a size 4, 7. For PSRL, $c = 0.75$. For f-Rmax, $m = 300, 500, 500, 500$ and for DORL, $c = 0.03$ in Circle 4, Circle 7, Three-leg 4, Three-leg 7, respectively.}
\label{fig:regret}
\end{figure}

There are two previously available sample efficient and implementable algorithms for FMDPs: factored $E^3$ and factored Rmax (f-Rmax). F-Rmax was shown to have better empirical performance \citep{guestrin2002algorithm}. Thus, we compare PSRL, DORL and f-Rmax. For PSRL, at the start of each episode, we simply sample each factored transition probability and reward functions from a Dirichlet distribution and a Gaussian distribution, i.e. $P_i^k(x) \sim \text{Dirichlet}(N^t_{P_i}(\cdot, x) / c)$ and $R_i^k(x) \sim N(\hat R_i^k(x), c/N^t_{P_i}(x))$, where $c$ is searched over $(0.05, 0.1, 0.3, 0.75, 1, 5, 20)$. The total number of samplings for PSRL in each round is upper bounded by the number of parameters of the FMDP. For DORL, we replace the coefficients 18 and 12 in (\ref{equ:delta_p}) and (\ref{equ:delta_r}) with a hyper-parameter $c$ searched over the set $\{0.05, 0.1, 0.3, 0.5, 1, 5, 20\}$. For f-Rmax, $m$, the number of visits needed to be known are chosen from 100, 300, 500, 700 and the best choice is selected for each experiment. 


For the approximate planner used by our algorithm, we implement approximate linear programming \citep{guestrin2003efficient} with the basis $h_i(s) = s_i$ for $i \in [m]$. For regret evaluation, we use an accurate planner to find the true optimal average reward.

We compare three algorithms on computer network administrator domain with a circle and a three-leg structure \citep{guestrin2001max,schuurmans2002direct}. To avoid the extreme case in our lower bound, both the MDPs are set to have limited diameters. The details on the environment are in Appendix \ref{app:exp_detail}.

Figure \ref{fig:regret} shows the regret of the two algorithms on circle and three-leg structure with a size 4, 7, respectively. Each experiment is run 20 times, with which median, 75\% and 25\% quantiles are computed. 
Our DORL and PSRL has very similar performance in all the environment except for Three-leg with a size 4. Optimal hyper-parameter for PSRL and DORL is stable in the way that $c$ around 0.75 and 0.03 are the optimal parameter for PSRL and DORL respectively for all the experiments. Note that we use the exact, not approximate, optimal reward in regret evaluation. So we see that DORL and PSRL was always able to find a near-optimal optimal policy despite the use of an approximate planner.

\section{Discussion}
In this paper, we provide two oracle-efficient algorithms PSRL and DORL for non-episodic FMDPs, with a Bayesian and frequentist regret bound of $\tilde{O}(D(l+m\sqrt{W})\sqrt{LT})$, respectively. PSRL outperforms previous near-optimal algorithm f-Rmax on computer network administration domain. The regret still converges despite using an approximate planner. We prove a lower bound of $\tilde{O}(\sqrt{sp(\vect{h}^+)LT})$ for non-episodic MDP. Our large diameter example shows that diameter $D$ can be arbitrary larger than the span $sp(\vect{h}^*)$. To reduce the gap, we propose factored span that scales with the difficulty of the graph structure and a new algorithm, FSRL with a regret bound of $\tilde{O}(Q\sqrt{WLT})$. In the lower bound construction, $Q$ equals to the span of the FMDP.

FSRL relies on a harder computational oracle that is not efficiently solvable yet. \citet{fruit2018efficient} achieved a regret bound depending on span using an implementable Bias-Span-Constrained value iteration on non-factored MDP. It remains unknown whether FSRL could be approximately solved using an efficient implementation.

In non-factored MDP, \cite{zhang2019regret} achieved the lower bound. On the lower bound of non-episodic FMDP, it remains an open problem to close  the remaining gap involving $\sqrt{W}$ and $\sqrt{Q}$.

Our algorithms require the full knowledge of the graph structure of the FMDP, which can be impractical. The structural learning scenario has been studied by \citet{strehl2007efficient,chakraborty2011structure,hallak2015off}. Their algorithms either rely on an admissible structure learner or do not have a regret or sample complexity guarantee. It remains an open problem of whether an efficient algorithm with theoretical guarantees exists for FMDP with unknown graph structure.

\newpage

\section{Broader Impacts}
As a theoretical paper, we can not foresee any direct societal consequences in the near future. Factored MDP, the main problem we study in this paper, may be used in multi-agent Reinforcement Learning scenario. 

\bibliography{main}

\begin{thebibliography}{}

\bibitem[Agrawal and Jia, 2017]{agrawal2017optimistic}
Agrawal, S. and Jia, R. (2017).
\newblock Optimistic posterior sampling for reinforcement learning: worst-case
  regret bounds.
\newblock In {\em Advances in Neural Information Processing Systems}, pages
  1184--1194.

\bibitem[Bartlett and Tewari, 2009]{bartlett2009regal}
Bartlett, P.~L. and Tewari, A. (2009).
\newblock Regal: A regularization based algorithm for reinforcement learning in
  weakly communicating mdps.
\newblock In {\em Proceedings of the Twenty-Fifth Conference on Uncertainty in
  Artificial Intelligence}, pages 35--42. AUAI Press.

\bibitem[Boutilier et~al., 2000]{boutilier2000stochastic}
Boutilier, C., Dearden, R., and Goldszmidt, M. (2000).
\newblock Stochastic dynamic programming with factored representations.
\newblock {\em Artificial intelligence}, 121(1-2):49--107.

\bibitem[Chakraborty and Stone, 2011]{chakraborty2011structure}
Chakraborty, D. and Stone, P. (2011).
\newblock Structure learning in ergodic factored mdps without knowledge of the
  transition function's in-degree.
\newblock In {\em Proceedings of the 28th International Conference on Machine
  Learning (ICML-11)}, pages 737--744. Citeseer.

\bibitem[Dann and Brunskill, 2015]{dann2015sample}
Dann, C. and Brunskill, E. (2015).
\newblock Sample complexity of episodic fixed-horizon reinforcement learning.
\newblock In {\em Advances in Neural Information Processing Systems}, pages
  2818--2826.

\bibitem[Dean and Kanazawa, 1989]{dean1989model}
Dean, T. and Kanazawa, K. (1989).
\newblock A model for reasoning about persistence and causation.
\newblock {\em Computational intelligence}, 5(2):142--150.

\bibitem[Fruit et~al., 2018]{fruit2018efficient}
Fruit, R., Pirotta, M., Lazaric, A., and Ortner, R. (2018).
\newblock Efficient bias-span-constrained exploration-exploitation in
  reinforcement learning.
\newblock In {\em International Conference on Machine Learning}, pages
  1578--1586.

\bibitem[Ghahramani, 1997]{ghahramani1997learning}
Ghahramani, Z. (1997).
\newblock Learning dynamic bayesian networks.
\newblock In {\em International School on Neural Networks, Initiated by IIASS
  and EMFCSC}, pages 168--197. Springer.

\bibitem[Givan et~al., 2000]{givan2000bounded}
Givan, R., Leach, S., and Dean, T. (2000).
\newblock Bounded-parameter markov decision processes.
\newblock {\em Artificial Intelligence}, 122(1-2):71--109.

\bibitem[Goldsmith et~al., 1997]{goldsmith1997complexity}
Goldsmith, J., Littman, M.~L., and Mundhenk, M. (1997).
\newblock The complexity of plan existence and evaluation in robabilistic
  domains.
\newblock In {\em Proceedings of the Thirteenth conference on Uncertainty in
  artificial intelligence}, pages 182--189.

\bibitem[Guestrin et~al., 2001]{guestrin2001max}
Guestrin, C., Koller, D., and Parr, R. (2001).
\newblock Max-norm projections for factored mdps.
\newblock In {\em IJCAI}, volume~1, pages 673--682.

\bibitem[Guestrin et~al., 2002a]{guestrin2002multiagent}
Guestrin, C., Koller, D., and Parr, R. (2002a).
\newblock Multiagent planning with factored mdps.
\newblock In {\em Advances in neural information processing systems}, pages
  1523--1530.

\bibitem[Guestrin et~al., 2003]{guestrin2003efficient}
Guestrin, C., Koller, D., Parr, R., and Venkataraman, S. (2003).
\newblock Efficient solution algorithms for factored mdps.
\newblock {\em Journal of Artificial Intelligence Research}, 19:399--468.

\bibitem[Guestrin et~al., 2002b]{guestrin2002algorithm}
Guestrin, C., Patrascu, R., and Schuurmans, D. (2002b).
\newblock Algorithm-directed exploration for model-based reinforcement learning
  in factored mdps.
\newblock In {\em ICML}, pages 235--242. Citeseer.

\bibitem[Guestrin et~al., 2002c]{guestrin2002context}
Guestrin, C., Venkataraman, S., and Koller, D. (2002c).
\newblock Context-specific multiagent coordination and planning with factored
  mdps.
\newblock In {\em AAAI/IAAI}, pages 253--259.

\bibitem[Hallak et~al., 2015]{hallak2015off}
Hallak, A., Schnitzler, F., Mann, T., and Mannor, S. (2015).
\newblock Off-policy model-based learning under unknown factored dynamics.
\newblock In {\em International Conference on Machine Learning}, pages
  711--719.

\bibitem[Jaksch et~al., 2010]{jaksch2010near}
Jaksch, T., Ortner, R., and Auer, P. (2010).
\newblock Near-optimal regret bounds for reinforcement learning.
\newblock {\em Journal of Machine Learning Research}, 11(Apr):1563--1600.

\bibitem[Kearns and Koller, 1999]{kearns1999efficient}
Kearns, M. and Koller, D. (1999).
\newblock Efficient reinforcement learning in factored mdps.
\newblock In {\em IJCAI}, volume~16, pages 740--747.

\bibitem[Kearns and Singh, 2002]{kearns2002near}
Kearns, M. and Singh, S. (2002).
\newblock Near-optimal reinforcement learning in polynomial time.
\newblock {\em Machine learning}, 49(2-3):209--232.

\bibitem[Kleinberg et~al., 2008]{kleinberg2008multi}
Kleinberg, R., Slivkins, A., and Upfal, E. (2008).
\newblock Multi-armed bandits in metric spaces.
\newblock In {\em Proceedings of the fortieth annual ACM symposium on Theory of
  computing}, pages 681--690. ACM.

\bibitem[Lewis and Puterman, 2001]{lewis2001probabilistic}
Lewis, M.~E. and Puterman, M.~L. (2001).
\newblock A probabilistic analysis of bias optimality in unichain markov
  decision processes.
\newblock {\em IEEE Transactions on Automatic Control}, 46(1):96--100.

\bibitem[Littman, 1997]{littman1997probabilistic}
Littman, M.~L. (1997).
\newblock Probabilistic propositional planning: representations and complexity.
\newblock In {\em Proceedings of the fourteenth national conference on
  artificial intelligence}, pages 748--754.

\bibitem[Luo et~al., 2018]{luo2018efficient}
Luo, H., Wei, C.-Y., Agarwal, A., and Langford, J. (2018).
\newblock Efficient contextual bandits in non-stationary worlds.
\newblock In {\em Conference On Learning Theory}, pages 1739--1776.

\bibitem[Osband et~al., 2013]{osband2013more}
Osband, I., Russo, D., and Van~Roy, B. (2013).
\newblock (more) efficient reinforcement learning via posterior sampling.
\newblock In {\em Advances in Neural Information Processing Systems}, pages
  3003--3011.

\bibitem[Osband and Van~Roy, 2014]{osband2014near}
Osband, I. and Van~Roy, B. (2014).
\newblock Near-optimal reinforcement learning in factored mdps.
\newblock In {\em Advances in Neural Information Processing Systems}, pages
  604--612.

\bibitem[Ouyang et~al., 2017]{ouyang2017learning}
Ouyang, Y., Gagrani, M., Nayyar, A., and Jain, R. (2017).
\newblock Learning unknown markov decision processes: A thompson sampling
  approach.
\newblock In {\em Advances in Neural Information Processing Systems}, pages
  1333--1342.

\bibitem[Puterman, 2014]{puterman2014markov}
Puterman, M.~L. (2014).
\newblock {\em Markov Decision Processes.: Discrete Stochastic Dynamic
  Programming}.
\newblock John Wiley \& Sons.

\bibitem[Schuurmans and Patrascu, 2002]{schuurmans2002direct}
Schuurmans, D. and Patrascu, R. (2002).
\newblock Direct value-approximation for factored mdps.
\newblock In {\em Advances in Neural Information Processing Systems}, pages
  1579--1586.

\bibitem[Strehl, 2007]{strehl2007model}
Strehl, A.~L. (2007).
\newblock Model-based reinforcement learning in factored-state mdps.
\newblock In {\em 2007 IEEE International Symposium on Approximate Dynamic
  Programming and Reinforcement Learning}, pages 103--110. IEEE.

\bibitem[Strehl et~al., 2007]{strehl2007efficient}
Strehl, A.~L., Diuk, C., and Littman, M.~L. (2007).
\newblock Efficient structure learning in factored-state mdps.
\newblock In {\em AAAI}, volume~7, pages 645--650.

\bibitem[Sun et~al., 2019]{sun2019model}
Sun, W., Jiang, N., Krishnamurthy, A., Agarwal, A., and Langford, J. (2019).
\newblock Model-based rl in contextual decision processes: Pac bounds and
  exponential improvements over model-free approaches.
\newblock In {\em Conference on Learning Theory}, pages 2898--2933.

\bibitem[Syrgkanis et~al., 2016]{syrgkanis2016improved}
Syrgkanis, V., Luo, H., Krishnamurthy, A., and Schapire, R.~E. (2016).
\newblock Improved regret bounds for oracle-based adversarial contextual
  bandits.
\newblock In {\em Advances in Neural Information Processing Systems}, pages
  3135--3143.

\bibitem[Tavakol and Brefeld, 2014]{tavakol2014factored}
Tavakol, M. and Brefeld, U. (2014).
\newblock Factored mdps for detecting topics of user sessions.
\newblock In {\em Proceedings of the 8th ACM Conference on Recommender
  Systems}, pages 33--40.

\bibitem[Zhang and Ji, 2019]{zhang2019regret}
Zhang, Z. and Ji, X. (2019).
\newblock Regret minimization for reinforcement learning by evaluating the
  optimal bias function.
\newblock In {\em Advances in Neural Information Processing Systems}, pages
  2823--2832.

\end{thebibliography}
\bibliographystyle{apalike}

\newpage
\onecolumn

\appendix

\section{Proof of Theorem \ref{thm:1} and \ref{thm:2}} \label{app:main_proof}

A standard regret analysis consists of proving the optimism, bounding the deviations and bounding the probability of failing the confidence set. Our analysis follows the standard procedure while adapting them to a FMDP setting. The novelty is on the proof of the general episode-assigning criterion and the lower bound.

\paragraph{Some notations.} For simplicity, we let $\pi^*$ denote the optimal policy of the true MDP, $\pi(M)$. Let $t_k$ be the starting time of episode $k$ and $K$ be the total number of episodes. Since $\tilde{R}^k(x, s)$ for any $(x, s) \in \tilde{\mathcal{X}}$ does not depend on $s$, we also let $\tilde{R}^k(x)$ denote $\tilde{R}^k(x, s)$ for any $s$. Let $\lambda^*$ and $\lambda_k$ denote the optimal average reward for $M$ and $M_k$.

\paragraph{Regret decomposition.}
We follow the standard regret analysis framework by \citet{jaksch2010near}. We first decompose the total regret into three parts in each episode:
\begin{align}
    R_T
    &= \sum_{t = 1}^T (\lambda^* - r_t)\nonumber\\
    &= \sum_{k=1}^{K}\sum_{t = t_k}^{t_{k+1}-1} (\lambda^* - \lambda_k)\label{equ:reg1} \\
    &+ \sum_{k=1}^{K}\sum_{t = t_k}^{t_{k+1}-1}(\lambda_k - R(s_t, a_t))\label{equ:reg2}\\
    &+ \sum_{k=1}^{K}\sum_{t = t_k}^{t_{k+1}-1}(R(s_t, a_t) - r_t). \label{equ:reg3}
\end{align}{}
Using Hoeffding's inequality, the regret caused by (\ref{equ:reg3}) can be upper bounded by $\sqrt{\frac{5}{2} T \log \left(\frac{8 }{\rho}\right)}$,  with probability at least $\frac{\rho}{12}$.

\paragraph{Confidence set.}
Let $\mathcal{M}_k$ be the confidence set of FMDPs at the start of episode $k$ with the same factorization, such that for and each $i\in[l]$,
$$
    |R_i(x) - \hat R^k_i(x)| \leq W_{R_i}^k(x), \forall x \in \mathcal{X}[Z_i^R],
$$
where $W_{R_i}^k(x)\coloneqq \sqrt{\frac{12\log(6l|\mathcal{X}[Z_i^R]|t_k/\rho)}{\max \{N_{R_i}^k(x), 1\} }}$ as defined in (\ref{equ:delta_r}); \\ and for each $j \in [m]$
$$
    |P_j(s|x) - \hat P^k_j(s|x)| \leq W_{P_j}^k(s|x), \forall x \in \mathcal{X}[Z_j^P], s \in \mathcal{S}_j,
$$
where $W_{P_j}^k(s|x)$ is defined in (\ref{equ:delta_p}). It can be shown that 
$$
    |P_j(x) - \hat P^k_j(x)|_1 
    \leq 2\sqrt{\frac{18|\mathcal{S}_i|\log(6S_im|\mathcal{X}[Z_i^P]|t_k/\rho)}{ \max \{N_{P_i}^k(x), 1\} }},
$$
where $\bar W_{P_i}^k(x) \coloneqq  2\sqrt{\frac{18|\mathcal{S}_i|\log(6S_im|\mathcal{X}[Z_i^P]|t_k/\rho)}{ \max \{N_{P_i}^k(x), 1\} }}$.

In the following analysis, we all assume that the true MDP $M$ for both PSRL and DORL are in $\mathcal{M}_k$ and $M_k$ by PSRL are in $\mathcal{M}_k$ for all $k \in [K]$. In the end, we will bound the regret caused by the failure of confidence set.

\subsection{Regret caused by difference in optimal gain}
We further bounded the regret caused by (\ref{equ:reg1}). For PSRL, since we use fixed episodes, we show that the expectation of (\ref{equ:reg1}) equals to zero.
\begin{lem}[Lemma 1 in \citet{osband2013more}]
If $\phi$ is the distribution of $M$, then, for any $\sigma(H_{t_k})-measurable$ function $g$,
$$
\mathbb{E}\left[g\left(M\right) | H_{t_{k}}\right]=\mathbb{E}\left[g\left(M_{k}\right) | H_{t_{k}}\right].
$$
\end{lem}{}
We let $g = \lambda(M, \pi(M))$. As $g$ is a $\sigma(H_{t_k})-measurable$ function. Since $t_k$, $K$ are fixed value for each $k$, we have $\mathbb{E}[\sum_{k=1}^{K} T_k (g(M) - g(M_k))] = 0$.

For DORL, we need to prove optimism, i.e, $\lambda(M_k, \tilde{\pi}_k) \geq \lambda^*$ with high probability. Given $M \in \mathcal{M}_k$, we show that there exists a policy for $M_k$ with an average reward $\geq \lambda^*$.

\begin{lem}
For any policy $\pi$ for $M$ and any vector $\vect{h} \in \mathbb{R}^{S}$, let $\tilde{\pi}$ be the policy for $M_k$ satisfying $\tilde{\pi}(s) = (\pi(s), s^*)$, where $s^* = \argmax_{s}\vect{h}(s)$. Then, given $M \in \mathcal{M}_k$, $(P( M_k, \tilde\pi) - P(M, \pi))\vect{h} \geq 0$.
\label{lem:opt}
\end{lem}



\begin{cor}
Let $\tilde{\pi}^*$ be the policy that satisfies $\tilde{\pi}^*(s) = (\pi^*(s), s^*)$, where $s^* = \max_s \vect{h}(M)$. Then $\lambda(M_k, \tilde\pi^*, s_1) \geq \lambda^*$ for any starting state $s_1$. \label{cor:opt}
\end{cor}{}

Proof of Lemma \ref{lem:opt} and Corollary \ref{cor:opt} are shown in Appendix \ref{app:opt}. Thereon, $\lambda(M_k, \tilde\pi_k) \geq \lambda(M_k, \tilde\pi^*, s_1) \geq \lambda^*$. The total regret of (\ref{equ:reg1}) $\leq 0$.
\subsection{Regret caused by deviation}
We further bound regret caused by (\ref{equ:reg2}), which can be decomposed into the deviation between our brief $M_k$ and the true MDP. We first show that the diameter of $M_k$ can be upper bounded by $D$.

\paragraph{Bounded diameter.} We need diameter of extended MDP to be upper bounded to give a sublinear regret. For PSRL, since prior distribution has no mass on MDP with diameter greater than $D$, the diameter of MDP from posterior is upper bounded by $D$ almost surely. For DORL, we have the following lemma, the proof of which is given in Appendix \ref{app:diameter}.

\begin{lem}
    When $M$ is in the confidence set $\mathcal{M}_k$, the diameter of the extended MDP $D(M_k) \leq D$.
    \label{lem:diameter}
\end{lem}

\paragraph{Deviation bound.}
Let $\nu_k(s, a)$ be the number of visits on $s, a$ in episode $k$ and $\vect{\nu}_k$ be the row vector of $\nu_k(\cdot, \pi_k(\cdot))$.
Let $\Delta_k = \sum_{s, a}\nu_k(s, a)(\lambda(M_k, \tilde{\pi}_k) - R(s, a))$. Using optimal equation,

\begin{align*}
 \Delta_k
    &= \sum_{s, a}\nu_k(s, a)\left[\lambda(M_k, \tilde \pi_k) - \tilde R^k(s, a)\right]\\
    &\quad + \sum_{s, a}\nu_k(s, a)\left[\tilde{R}^k(s, a) - R(s, a) \right]\\
    &= \vect{\nu}_k(\tilde P^k - I) \vect{h}_k + \vect{\nu}_k (\tilde{\vect{R}}^k - \vect{R}^k)\\
    &=  \underbrace{\vect{\nu}_k(P^k - I)\vect{h}_k}_{\textcircled{1}} + \underbrace{\vect{\nu}_k(\tilde P^k - P^k)\vect{h}_k}_{\textcircled{2}} + \underbrace{\vect{\nu}_k (\tilde{\vect{R}}^k - \vect{R}^k)}_{\textcircled{3}},
\end{align*}
where $\tilde{P}^k \coloneqq P(M_k, \tilde{\pi}_k), P^k \coloneqq P(M, \pi_k), \vect{h}_k \coloneqq \vect{h}^*(M_k)$,  and $\tilde{\vect{R}}^k \coloneqq \vect{R}(M_k, \tilde{\pi}_k), \vect{R}^k \coloneqq \vect{R}(M, \pi_k)$.

%

Using Azuma-Hoeffding inequality and the same analysis in \cite{jaksch2010near}, we bound $\textcircled{1}$ with probability at least $1 - \frac{\rho}{12}$,
\begin{equation}
    \sum_k\textcircled{1} = \sum_k\vect{\nu}_{k}\left(P^{k}-I\right) \vect{h}_{k} \leq D \sqrt{\frac{5}{2} T \log \left(\frac{8}{\rho}\right)} + KD. \label{equ:K}
\end{equation}



To bound $\textcircled{2}$ and $\textcircled{3}$, we analyze the deviation in transition and reward function between $M$ and $M_k$. For DORL, the deviation in transition probability is upper bounded by
\begin{align*}
&\quad \max_{s^{\prime}}|\tilde{P}^k_i(x, s^{\prime}) - \hat{P}^k_i(x)|_1 \\
    &\leq \min\{2\sum_{s\in\mathcal{S}_i}W_{P_i}^k(s \mid x), 1\}\\
    &\leq \min\{2\bar W_{P_i}^k(x), 1\} \leq 2\bar W_{P_i}^k(x),
\end{align*}
The deviation in reward function $|\tilde{R}_i^k - \hat R_i^k|(x) \leq W_{R_i}^k(x)$.

For PSRL, since $M_k \in \mathcal{M}_k$, $|\tilde{P}_i^k - \hat P_i^k|(x) \leq \bar W_{P_i}^k(x)$ and $|\tilde{R}_i^k - \hat  R_i^k|(x) \leq W_{R_i}^k(x)$.

Decomposing the bound for each scope provided by $M \in \mathcal{M}_k$ and $M_k$ for PSRL $\in \mathcal{M}_k$, it holds for both PSRL and DORL that:
\begin{align}
    \sum_k\textcircled{2} 
    &\leq 3\sum_k D\sum_{i=1}^m\sum_{x \in \mathcal{X}[Z_i^P]} \nu_k(x)\bar W_{P_i}^k(x), \label{equ:p_dev}\\
    \sum_k\textcircled{3} 
    &\leq 2\sum_k\sum_{i=1}^l\sum_{x \in \mathcal{X}[Z_i^R]} \nu_k(x)W_{R_i}^k(x); \label{equ:r_dev}
\end{align}
where with some abuse of notations, define $\nu_k(x) = \sum_{x^{\prime} \in \mathcal{X}: x^{\prime
}[Z_i] = x} \nu_k(x^{\prime})$ for $x \in \mathcal{X}[Z_i]$. The second inequality is from the fact that $|\tilde P^k(\cdot | x) - P^k(\cdot | x)|_1 \leq \sum_1^m |\tilde P^k_i(\cdot | x[Z_i^R]) - P^k_i(\cdot | x[Z_i^R])|_1$ \citep{osband2014near}.

\subsection{Balance episode length and episode number}
We give a general criterion for bounding (\ref{equ:K}), (\ref{equ:p_dev}) and (\ref{equ:r_dev}).

\begin{lem}
    For any fixed episodes $\{T_k\}_{k = 1}^K$, if there exists an upper bound $\bar T$, such that $T_k \leq \bar T$ for all $k \in [K]$, we have the bound 
    $$
    \sum_{x\in \mathcal{X}[Z]}\sum_k \nu_k(x)/\sqrt{\max\{1, N_k(x)\}} \leq L\bar T + \sqrt{LT},
    $$
    where $Z$ is any scope with $|\mathcal{X}[Z]| \leq L$, and $\nu_k(x)$ and $N_k(x)$ are the number of visits to $x$ in and before episode $k$. Furthermore, total regret of  (\ref{equ:K}), (\ref{equ:p_dev}) and (\ref{equ:r_dev}) can be bounded by $\tilde{O}\big((\sqrt{W}Dm+l)(L\bar T + \sqrt{LT}) + KD\big)$ \label{lem:epi_cri}
\end{lem}{}

Lemma \ref{lem:epi_cri} implies that bounding the deviation regret is to balance total number of episodes and the length of the longest episode. The proof, as shown in Appendix \ref{app:epi_cri}, relies on defining the last episode $k_0$, such that $N_{k_0}(x) \leq \nu_{k_0}(x)$.

Instead of using the doubling trick that was used in \cite{jaksch2010near}. We use an arithmetic progression: $T_k = \lceil k/L \rceil$ for $k \geq 1$. 
As in our algorithm, $T \geq \sum_{k = 1}^{K-1}T_k \geq \sum_{k = 1}^{K-1}k/L = \frac{(K-1)K}{2L}$, we have $K \leq \sqrt{3LT}$ and $T_k \leq T_K \leq K/L \leq \sqrt{3T/L}$ for all $k \in [K]$. Thus, by Lemma \ref{lem:epi_cri}, putting (\ref{equ:reg3}), (\ref{equ:K}), (\ref{equ:r_dev}), (\ref{equ:p_dev}) together, the total regret for $M \in \mathcal{M}_k$ is upper bounded by
\begin{align*}
    \tilde{O}\big((\sqrt{W}Dm + l)\sqrt{LT} \big), \numberthis \label{equ:reg_sum1}
\end{align*}
with a probability at least $1 - \frac{\rho}{6}$.

For the failure of confidence set, we prove the following Lemma in Appendix \ref{app:reg_failure}.
\begin{lem}
    For all $k \in [K]$, with probability greater than $1 - \frac{3\rho}{8}$, $M \in \mathcal{M}_k$ holds.
\end{lem}{}
Combined with (\ref{equ:reg_sum1}), with probability at least $1-\frac{2\rho}{3}$ the regret bound in Theorem \ref{thm:2} holds.

For PSRL, $M_k$ and $M$ has the same posterior distribution. The expectation of the regret caused by $M \notin \mathcal{M}_k$ and $M_k \notin \mathcal{M}_k$ are the same. Choosing sufficiently small $\rho \leq \sqrt{1/T}$, Theorem \ref{thm:1} follows.

\section{Optimism (Proof of Lemma \ref{lem:opt} and Corollary \ref{cor:opt})} \label{app:opt}

\begin{lem}
For any policy $\pi$ for $M$ and any vector $\vect{h} \in \mathbb{R}^{S}$, let $\tilde{\pi}$ be the policy for $M_k$ satisfying $\tilde{\pi}(s) = (\pi(s), s^*)$, where $s^* = \argmax_{s}\vect{h}(s)$. Then, given $M \in \mathcal{M}_k$, $(P( M_k, \tilde\pi) - P(M, \pi))\vect{h} \geq 0$.
\label{lem:opt_appendix}
\end{lem}

\textit{Proof.} We fix some $s \in \mathcal{S}$ and let $x = (s, \pi(s)) \in \mathcal{X}$. Recall that for any $s_i \in \mathcal{S}_i$, $\Delta_i^k(s_i|x) = $
$$
\min \{\sqrt{\frac{18 \hat{P}_{i}^{k}(s_i | x) \log \left(c_{i, k}\right)}{\max \left\{N_{P_{i}}^{k}(x), 1\right\}}} + \left.\frac{18 \log \left(c_{i, k}\right)}{\max \left\{N_{P_{i}}^{k}(x), 1\right\}}, \hat{P}_{i}^{k}(s_i | x)\right\}.
$$
and define $P^{-}_i(\cdot | x) = \hat P^k_i(\cdot | x) - \Delta^k_i(\cdot | x)$. Slightly abusing the notations, let
$\tilde{\vect{P}} = P(M_k, \tilde{\pi})_{s, \cdot}$,
$\vect{P} = P(M, \pi)_{s, \cdot}$. Define two $S$-dimensional vectors $\hat{\vect{P}}$ and $\vect{P}^{-}$ with
$\hat{\vect{P}}(\bar s) = \prod_i\hat{P}_i(\bar s[Z_i^P]|x)$ and $\vect{P}^{-}(\bar s) = \Pi_{i}P_i^{-}(\bar s[Z_i^P] | x)$ for $\bar s \in \mathcal{S}$.

As $M \in \mathcal{M}_k$, $\vect{P}^{-} \leq \vect{P}$. Define $\vect{\alpha} \coloneqq \hat{\vect{P}} - \vect{P} \leq \hat{\vect{P}} - \vect{P}^{-} \eqqcolon \vect{\Delta}$. Without loss of generality, we let $\max_s \vect{h}(s) = D$.

$$
\begin{aligned}
\sum_{i} \tilde{\vect{P}}(i) \vect{h}(i)
&=\sum_{i} \vect{P}(i)^{-} \vect{h}(i)+D\left(1-\sum_{j} \vect{P}(j)^{-}\right)\\
&=\sum_{i} \vect{P}(i)^{-} \vect{h}(i)+D \sum_{j} \vect{\Delta}(j) \\
&=\sum_{i}\left(\hat{\vect{P}}(i)-\vect{\Delta}(i)\right) \vect{h}(i)+D \vect{\Delta}(i)\\
&=\sum_{i} \hat{\vect{P}}(i) \vect{h}(i)+\left(D-\vect{h}(i)\right) \vect{\Delta}(i) \\
& \geq \sum_{i} \hat{\vect{P}}(i) \vect{h}(i)+\left(D-\vect{h}(i)\right) \vect{\alpha}(i)\\
&=\sum_{i}\left(\hat{\vect{P}}(i)-\vect{\alpha}(i)\right) \vect{h}(i)+D \vect{\alpha}(i) \\
&=\sum_{i} \vect{P}(i) \vect{h}(i)+D \sum_{i} \vect{\alpha}(i)=\sum_{i} \vect{P}(i) \vect{h}(i) \end{aligned}
$$

\begin{cor}
Let $\tilde{\pi}^*$ be the policy that satisfies $\tilde{\pi}^*(s) = (\pi^*(s), s^*)$, where $s^* = \max_s \vect{h}(M)$. Then $\lambda(M_k, \tilde\pi^*, s_1) \geq \lambda^*$ for any starting state $s_1$.
\end{cor}{}
\textit{Proof.} Let $\vect{d}(s_1) \coloneqq \vect{d}(M_k, \tilde\pi^*, s_1) \in \mathbb{R}^{1\times S}$ be the row vector of stationary distribution starting from some $s_1 \in \mathcal{S}$.
By optimal equation,
\begin{align*}
    &\quad\lambda( M_k, \tilde\pi^*, s_1) - \lambda^*\\
    &= \vect{d}(s_1) \vect{R}( M_k, \tilde\pi^*) -  \lambda^*(\vect{d}(s_1) \mathbf{1})\\
    &= \vect{d}(s_1)(\vect{R}( M_k, \tilde\pi^*) - \lambda^*\mathbf{1})\\
    &= \vect{d}(s_1)(\vect{R}( M_k, \tilde\pi^*) - \vect{R}(M, \pi^*)) \\
    &\quad + \vect{d}(s_1)(I - P(M, \pi^*))\vect{h}(M)\\
    &\geq \vect{d}(s_1)(\vect{R}( M_k, \tilde\pi^*) - \vect{R}(M, \pi^*)) \\
    &\quad + \vect{d}(s_1)(P( M_k, \tilde\pi^*) - P(M, \pi^*))\vect{h}(M)\\
    &\geq 0,
\end{align*}
where the last inequality is by Lemma \ref{lem:opt} and Corollary \ref{cor:opt} follows.


\section{Proof of Lemma \ref{lem:diameter}} \label{app:diameter}
\begin{lem}
    Given $M$ in the confidence set $\mathcal{M}_k$, the diameter of the extended MDP $D(M_k) \leq D$.
\end{lem}

\textit{Proof.} Fix a $s_1 \neq s_2$, there exist a policy $\pi$ for $M$ such that the expected time to reach $s_2$ from $s_1$ is at most $D$, without loss of generality we assume $s_2$ is the last state. Let $E$ be the $(S-1) \times 1$ vector with each element to be the expected time to reach $s_2$ except for itself. We find $\tilde \pi$ for $M_k$ such that the expected time to reach $s_2$ from $s_1$ can be bounded by $D$. We choose the $\tilde \pi$ that satisfies $\tilde{\pi}(s) = (\pi(s), s_2)$.


Let $Q$ be the transition matrix under $\tilde \pi$ for $M_k$. Let $Q^{-}$ be the matrix removing $s_2$-th row and column and $P^{-}$ defined in the same way for $M$. We immediately have $P^{-1}E \geq Q^{-1}E$, given $M \in \mathcal{M}_k$. Let $\tilde E$ be the expected time to reach $s_2$ from every other states except for itself under $\tilde \pi$ for $M_k$.

We have $\tilde E = \vect{1} + Q^{-}\tilde E$. The equation for $E$ gives us $E = \vect{1} + P^{-}E \geq \vect{1} + Q^{-}E$. Therefore,
$$
    \tilde E = (1-Q^{-})^{-1}\vect{1} \leq E,
$$
and $\tilde E_{s_1} \leq E_{s_1} \leq D$. Thus, $D(M_k) \leq D$.

\section{Deviation bound (Proof of Lemma \ref{lem:epi_cri})}\label{app:epi_cri}

\begin{lem}
    For any fixed episodes $\{T_k\}_{k = 1}^K$, if there exists an upper bound $\bar T$, such that $T_k \leq \bar T$ for all $k \in [K]$, we have the bound 
    $$
    \sum_{x\in \mathcal{X}[Z]}\sum_k \nu_k(x)/\sqrt{\max\{1, N_k(x)\}} \leq L\bar T + \sqrt{LT},
    $$
    where $Z$ is any scope, and $\nu_k(x)$ and $N_k(x)$ are the number of visits to $x$ in and before episode $k$. Furthermore, the total regret of  (\ref{equ:K}), (\ref{equ:p_dev}) and (\ref{equ:r_dev}) can be bounded by $(\sqrt{W}Dm+l)(L\bar T + \sqrt{LT}) + KD$.
\end{lem}{}


\textit{Proof.} We bound the random variable $\sum_{k = 1}^{K} \frac{\nu_k(x)}{\sqrt{\max\{N_k(x), 1\}}}$ for every $x \in \mathcal{X}[Z]$, where $\nu_k(x) = \sum_{t = t_k}^{t_{k+1} - 1}\mathbbm{1}(x_t = x)$ and $N_k(x) = \sum_{i=1}^{k-1}\nu_k(x)$.

Let $k_0(x)$ be the largest $k$ such that $N_k(x) \leq \nu_k(x)$. Thus $\forall k \geq k_0(x), N_k(x) > \nu_k(x)$, which gives $N_t(x) \coloneqq N_k(x) + \sum_{\tau = t_k}^{t}\mathbbm{1}(x_{\tau} = x) < 2N_k(x)$ for $t_k \leq t < t_{k+1}$.

Conditioning on $k_0(x)$, we have 
\begin{align*}
    &\quad \sum_{k = 1}^{K} \frac{\nu_k(x)}{\sqrt{\max\{N_k(x), 1\}}}\\
    &\leq N_{k_0(x)}(x) + \nu_{k_0(x)}(x) + \sum_{k > k_0(x)} \frac{\nu_k(x)}{\sqrt{\max\{N_k(x), 1\}}}\\
    &\leq 2\nu_{k_0(x)}(x) + \sum_{k > k_0(x)} \frac{\nu_k(x)}{\sqrt{\max\{N_k(x), 1\}}}\\
    &\leq 2\bar T + \sum_{k > k_0(x)} \frac{\nu_k(x)}{\sqrt{\max\{N_k(x), 1\}}},
\end{align*}{}
where the first inequality uses $\max\{N_k(x), 1\} \geq 1$ for $k = 1, \dots k_0(x)$, the second inequality is by the fact that $N_{k_0(x)}(x) \leq \nu_{k_0(x)}(x)$ and the third one is by $\nu_{k_0}(x) \leq T_{k_0(x)} \leq T_K$.

And letting $k_1(x) = k_0(x) + 1$ and $N(x) \coloneqq N_K(x) + \nu_K(x)$, we have
\begin{align*}
    &\quad \sum_{k > k_0(x)} \frac{\nu_k(x)}{\sqrt{\max\{N_k(x), 1\}}}\\
    &\leq \sum_{t = t_{k_1(x)}}^{T} 2\frac{\mathbbm{1}(x_t = x)}{\sqrt{\max\{N_t(x), 1\}}}\\
    &\leq \sum_{t = t_{k_1(x)}}^{T} 2\frac{\mathbbm{1}(x_t = x)}{\sqrt{\max\{N_t(x)-N_{k_1(x)}, 1\}}}\\
    &\leq 2 \int_{1}^{N(x) - N_{k_1(x)}} \frac{1}{\sqrt{x}} dx\\
    &\leq (2+\sqrt{2})\sqrt{N(x)}.
\end{align*}{}

Given any $k_0(x)$, we can bound the term with a fixed value $2\bar T + (2+\sqrt{2})\sqrt{N(x)}$. Thus, the random variable $\sum_{k = 1}^{K} \frac{\nu_k(x)}{\sqrt{\max\{N_k(x), 1\}}}$ is upper bounded by $ 2\bar T + (2+\sqrt{2})\sqrt{N(x)}$ almost surely. Finally, $\sum_x\sum_{k = 1}^{K} \frac{\nu_k(x)}{\sqrt{\max\{N_k(x), 1\}}} \leq L\bar T + (2+\sqrt{2})\sqrt{LT}$. The regret by (\ref{equ:p_dev}) is 
\begin{align*}
    \sum_{k}3D\sum_{i \in [m]}\sum_{x \in \mathcal{X}[Z_i^P]}\nu_k(x)\bar W^k_{P_i}(x) \\= \tilde{O}(\sqrt{W}Dm(L\bar T + \sqrt{LT}) + KD).
\end{align*}{}

The regret by (\ref{equ:r_dev}) is
$$
    \sum_{k} 2\sum_{i \in [l]}\sum_{x \in \mathcal{X}[Z_i^R]}\nu_k(x)\bar W^k_{R_i}(x) = \tilde{O}(l(L\bar T + \sqrt{LT}) + KD).
$$
The last statement is completed by directly summing (\ref{equ:K}), (\ref{equ:p_dev}) and (\ref{equ:r_dev}).
\section{Regret caused by failing confidence bound}\label{app:reg_failure}

\begin{lem}
    For all $k \in [K]$, with probability greater than $1 - \frac{3\rho}{8}$, $M \in \mathcal{M}_k$ holds.
\end{lem}{}

\textit{Proof.} We first deal with the probabilities, with which in each round a reward function of the true MDP $M$ is not in the confidence set.
Using Hoeffding's inequality, we have for any $t, i$ and $x \in \mathcal{X}[Z_i^R]$,
\begin{align*}
    &\quad \mathbbm{P}\left\{ |\hat{R}^t_i(x) - R_i(x)| \geq \sqrt{\frac{12\log(6l|\mathcal{X}[Z_i^R]| t/\rho)}{\max\{1, N^t_{R_i}(x)\}}} \right\} \\
    &\leq  \frac{\rho}{3l|\mathcal{X}[Z_i^R]|t^6}, \text{ with a summation} \leq \frac{3}{12}\rho.
\end{align*} Thus, with probability at least $1-\frac{3\rho}{12}$, the true reward function is in the confidence set for every $t \leq T$.

For the transition probability, we use a different concentration inequality.
\begin{lem}[Multiplicative Chernoff Bound \citep{kleinberg2008multi} Lemma 4.9]
    Consider $n$, i.i.d random variables $X_1, \dots, X_n$ on $[0, 1]$. Let $\mu$ be their mean and let $X$ be their average. Then with probability $1-\rho$,
    $$
        |X-\mu| \leq \sqrt{\frac{3 \log (2 / \rho) X}{n}}+\frac{3 \log (2 / \rho)}{n}.
    $$ \label{lem:chernoff}
\end{lem}{}

Using Lemma \ref{lem:chernoff}, for each $x, i, k$, it holds that with probability $1 - \rho/(6m\left|\mathcal{X}\left[Z_{i}^{P}\right]\right| t_{k}^6)$,
$$
    |\hat P_i(\cdot | x) - P_i(\cdot | x)|_1 \leq \sqrt{\frac{18 S_i \log (c_{i, k})}{\max \{N_{P_i}^k(x), 1\}}}+\frac{18 \log (c_{i, k})}{\max \{N_{P_i}^k(x), 1\}}.
$$
Then with a probability $1-\frac{3\rho}{24}$, it holds for all $x, i, k$. Therefore, with a probability $1-\frac{3\rho}{8}$, the true MDP is in the confidence set for each $k$.

\section{Span of Cartesian product of MDPs} \label{app:prod}
\begin{lem}
    Let $M^{+}$ be the Cartesian product of $n$ independent MDPs $\{M_i\}_{i = 1}^n$, each with a span of bias vector $sp(h_i)$. The optimal policy for $M^+$ has a span $sp(h^+) = \sum_i sp(h_i)$.
\end{lem}

\textit{Proof.} Let $\lambda^*_i$ for $i \in [n]$ be the optimal gain of each MDP. Optimal gain of $M^+$ is direct $\lambda^* = \sum_{i \in [n]} \lambda^*_i$. As noted in \cite{puterman2014markov} (8.2.3), by the definition of bias vector we have 
$$
    h_i(s) = \mathbb{E}[\sum_{t = 1}^{\infty} (r^i_t - \lambda^*_i)\mid s^i_1 = s], \quad \forall s \in \mathcal{S}_i,
$$
where $r_t^i$ is the reward of the $i$-th MDP at time $t$ and $s^i_t \coloneqq s_t[i]$.

The lemma is directly by
\begin{align*}
h^+(s) &= \mathbb{E}[\sum_{t = 1}^{\infty} (r_t - \lambda^*)\mid s_1 = s]\\
&= \mathbb{E}[\sum_{t = 1}^\infty (\sum_{i \in [n]} (r_t^i - \lambda_i^*)) \mid s_1 = s]\\
&= \sum_{i \in [n]}\mathbb{E}[\sum_{t = 1}^\infty (r_t^i - \lambda_i^*) \mid s^i_1 = s[i]]\\
&= \sum_{i \in [n]} h_i(s[i]).
\end{align*}{}

We immediately have $sp(h^+) = \sum_i sp(h_i)$.

\section{Proof of Theorem \ref{thm:4}} \label{app:thm4}
{\it Proof.} Starting from $\textcircled{2}$, for each $s \in \mathcal{S}$, we bound $(\tilde P^k(\cdot \mid s) - P^k(\cdot \mid s))\vect{h}_k$. For simplicity, we remove the subscriptions of $s$ and use $\tilde P^k$ and $P^k$ to denote the vector for $s$-th row of the two matrix.
\begin{align*}
    &\quad \sum_{s \in \mathcal{S}}(\tilde P^k(s) - P^k(s)) h_k(s)\\
    &= \sum_{s_1 \in \mathcal{S}_1} \sum_{s_{-1} \in \mathcal{S}^{-1}} (P_1(s_1)P_{-1}(s_{-1}) - \tilde{P}_1(s_1) \tilde{P}_{-1}(s_{-1})) h_k(s_1, s_{-1})\\
    &= \sum_{s_1} \left[(P_1(s_1) - \tilde{P}_1(s_1)) \sum_{s_{-1}} \tilde{P}_{-1}(s_{-1}) h_k(s_1, s_{-1})\right] + \\
     &\quad \sum_{s_{-1}} \left[(P_{-1}(s_{-1}) - \tilde{P}_{-1}(s_{-1})) \sum_{s_{}} P_{1}(s_{1}) h_k(s_1, s_{-1})\right] \\
    &= \sum_{s_1}(P_1(s_1) - \tilde{P}_{1}(s_1)) h_{1k}(s_1) + \sum_{s_{-1}}(P_{-1}(s_{-1}) - \tilde{P}_{-1}(s_{-1})) h_{-1k}(s_{-1}),
\end{align*}{}
where $h_{1k}(s_1) \coloneqq \sum_{s_{-1}} \tilde{P}_{-1}(s_{-1}) h_k(s_1, s_{-1})$ and $h_{-1k}(s_{-1}) \coloneqq \sum_{s_{1}} P_{1}(s_{1}) h_k(s_1, s_{-1})$. As $span(h_{1k}) \leq sp_1(M_k)$, 
\begin{equation}
    \sum_{s \in \mathcal{S}}(\tilde P^k(s) - P^k(s)) h_k(s) \leq |P_1 - \tilde{P}_1|_1 sp_1(M_k) + \sum_{s_{-1}}(P_{-1}(s_{-1}) - \tilde{P}_{-1}(s_{-1})) h_{-1k}(s_{-1}).
    \label{equ:induction}
\end{equation}
By applying (\ref{equ:induction}) recurrently, we have 
$$
    \sum_{s \in \mathcal{S}}(\tilde P^k(s) - P^k(s)) h_k(s) \leq \sum_{i = 1}^m|P_i - \tilde{P}_i|_1 sp_i(M_k).
$$
Note that $sp_i(M_k)$ is generally smaller than $span(h_k)$. In our lower bound case each $sp_i = \frac{1}{m}span(h_k)$, which improves our upper bound by a scale of $1/m$.

The reduction of $l$ can be achieved by bounding each factored reward to be in $[1, 1/l]$. The following proof remains the same.



\section{Proof of Lower Bound} \label{app:lowerbound}
\textit{Proof sketch.} Let $l = |\cup_i^{n} Z_i^R|$. As $i \in Z_i^P$, a special case is the FMDP with graph structure $\mathcal{G} = \left(\left\{\mathcal{S}_{i}\right\}_{i=1}^{n} ;\left\{\mathcal{S}_i \times \mathcal{A}_{i}\right\}_{i=1}^{n} ;\left\{\{i\}\right\}_{i=1}^{l} \text{and} \left\{\emptyset\right\}_{i = l+1}^{n} ;\left\{\{i\}\right\}_{i=1}^{n}\right)$, which can be decomposed into $n$ independent MDPs as in the previous example. Among the $n$ MDPs, the last $n-l$ MDPs are trivial. By simply setting the rest $l$ MDPs to be the construction used by \cite{jaksch2010near}, which we refer to as "JAO MDP", the regret for each MDP with the span $sp(\vect{h})$, is $\Omega(\sqrt{sp({\vect{h}})WT})$ for $i \in [l]$. The total regret is $\Omega(l\sqrt{sp({\vect{h}})WT})$. 

\begin{lem}
    Let $M^{+}$ be the Cartesian product of $n$ independent MDPs $\{M_i\}_{i = 1}^n$, each with a span of bias vector $sp(h_i)$. The optimal policy for $M^+$ has a span $sp(h^+) = \sum_i sp(h_i)$. \label{lem:prod}
\end{lem}{}

Using Lemma \ref{lem:prod} (proved in Appendix \ref{app:prod}), $sp(\vect{h}^+) =  l\ sp({\vect{h}})$ and the total expected regret is $\Omega(\sqrt{l\ sp(\vect{h}^+)WT})$. Normalizing the reward function to be in $[0, 1]$, the expected regret of the FMDP is $\Omega(\sqrt{sp(\vect{h}^+)WT})$, which completes the proof.

\section{FSRL algorithm}
\label{sec:fsrl}

Here we provide a complete description of the FSRL algorithm that was omitted in the main paper due to space considerations.

\begin{algorithm}[H]
\caption{FSRL}
\begin{algorithmic}
   \State {\bfseries Input:} $\mathcal{S}, \mathcal{A}$, $T$, encoding $\mathcal{G}$ and upper bound on sum of factored span $Q$.
   \State $k \leftarrow 1; t \leftarrow 1; t_k \leftarrow 1; T_k = 1; \mathcal{H} \leftarrow \{\}$
   \Repeat
   \State Choose $M_k \in \mathcal{M}_k$ by solving the following optimization over $M \in \mathcal{M}_k$,
   $$
    \max \lambda^*(M) \quad \text{subject to} \quad Q(h) \leq Q \text{ for $h$ being the bias vector of $M$}. 
   $$
   \State Compute $\tilde \pi_k = \pi(M_k) 
  $.
   \For{$t=t_k$ {\bfseries to} $t_k + T_k - 1$}
   \State Apply action $a_t = \pi_k(s_t)$
   \State Observe new state $s_{t+1}$
   \State Observe new rewards $r_{t+1} = (r_{t+1, 1}, \dots r_{t+1, l})$
   \State $\mathcal{H} = \mathcal{H} \cup \{(s_t, a_t, r_{t+1}, s_{t+1})\}$
   \State $t \leftarrow t + 1$
   \EndFor
   \State $k \leftarrow k + 1$.
   \State $T_{k} \leftarrow \lceil k/L \rceil$; $t_k \leftarrow t+1$.
   \Until{$t_k > T$}
\end{algorithmic}
\label{alg:fregal}
\end{algorithm}

\end{document}